# Deep Learning for Bridge Load Capacity Estimation in Post-Disaster and -Conflict Zones

Arya Pamuncak[1], Weisi Guo[1,2], Ahmed Soliman Khaled[1], Irwanda Laory[1]

[1]*School of Engineering, University of Warwick, Coventry, United Kingdom*
[2]*Alan Turing Institute, London, United Kingdom*

## 1. Abstract

Many post-disaster and -conflict regions do not have sufficient data on their transportation infrastructure assets, hindering both mobility and reconstruction. In particular, as the number of aging and deteriorating bridges increase, it is necessary to quantify their load characteristics in order to inform maintenance and prevent failure. The load carrying capacity and the design load are considered as the main aspects of any civil structures. Human examination can be costly and slow when expertise is lacking in challenging scenarios. In this paper, we propose to employ deep learning as method to estimate the load carrying capacity from crowd sourced images. A new convolutional neural network architecture is trained on data from over 6000 bridges, which will benefit future research and applications. We tackle significant variations in the dataset (e.g. class interval, image completion, image colour) and quantify their impact on the prediction accuracy, precision, recall and F1 score. Finally, practical optimisation is performed by converting multiclass classification into binary classification to achieve a promising field use performance.

## 2. Introduction

### 2.1. Bridge Load Testing

Whilst government agencies typically maintain a database of bridge and infrastructure status, post-disaster and post-conflict scenarios can often lead to the loss of this knowledge. As many of our bridges are aging and deteriorating, it is essential to understand the structures condition to improve maintenance efficiency and avoid catastrophic events. In assessing a bridge's safety, the load level that can be applied to the bridge during its life time should be regarded as one of the most important properties in bridge maintenance since the presence of overweight vehicles can lead to decrease in the service life [1]. According to American Association of State Highway and Transportation Officials (AASHTO) *"Manual for condition evaluation and load and resistance factor rating (LRFR) of highway bridges",* load rating is generally used as a parameter which defines the load carrying capacity of a bridge [2]. In obtaining load rating of a bridge, visual inspection and load testing are commonly performed [3]. Whilst visual inspection is rapid, truck load tests are common to remove subjective bias. However, the test requires a pre-weighed truck, bridge instrumentation, and bridge closure which can be slow and expensive. In addition to load rating, design load can be utilised as another parameter which represent a bridge's load carrying capacity. According to [2], design load describes the assumption of live load used when a bridge is designed. Therefore, the bridge condition will not affect the design load since this load level is only employed in the design phase of a bridge. In [4] the design load has been classified into 13 classes and this class represent a load level that is used in the design step of a bridge in America. There are clear limitations related to the aforementioned activities such as availability of resources, subjectivity, cost, and time. This is exasperated in post-disaster zones, where expertise is expensive and sparse, and the need to rapidly understand the capacity of bridges is critical for recovery.

### 2.2. Deep Learning

*Arya Pamuncak (A.Pamuncak@warwick.ac.uk).
†Present address: School of Engineering, University of Warwick

During the past decade, scalable machine learning techniques that can process large amounts of high-dimensional data has transformed many sectors, such as automated driving, medical imaging, and natural language processing. In particular, deep learning methods of replaced conventional feature design-based machine learning by automating the feature extraction process and reducing human domain expertise [5, 8]. Deep learning achieves this through many layers of processing stages hierarchically in performing its task [5-7]. Currently, deep learning has gained popularity in the image processing field, particularly due to the rise of Convolutional Neural Networks (CNN). CNN has architectures which are inspired by part of mammalian brain which is known as the primary visual cortex in processing visual input [7] and it can be seen as the first successful deep learning architecture [5, 6]. Introduced in the early 1990's, CNNs rose into prominence in 2012 when it produced the best performance in ImageNet competition. Following this success, CNN has revolutionised the field of computer vision and become a state of the art for recognition and detection purpose [5]. In addition, it has been utilised in a wide range of application such as image classification for a large number of classes [9-11], traffic sign recognition [12], medical object classification [13-15], face recognition [16, 17], and damage detection in structures [18-21]. In addition to automatic feature extraction, CNN can produce excellent performance for complex image recognition task due to the capability of CNN in exploiting the local spatial correlation between pixels in the image [5]. Unlike other image recognition algorithm, CNN depends on the spatial separation instead of depending the spatial position hence combination of local features is more important than the location of features which might be varied on images. This promising achievement and the ubiquitous use of mobile camera have motivated the research in the development of a deep learning-based method for load rating estimation using image.

### 2.3. Machine Learning in Civil Engineering Maintenance

Some efforts have been made in automatic defect detection by utilising CNNs in image processing. In 2017, Young-Jin et al. employed CNN technique for crack detection on concrete surfaces [18]. In this research, 40,000 images are used to train a neural network and 55 testing images are utilised to observe the network performance. Comparative study with traditional Canny and Sobel edge detection method is conducted to observe the performance of the proposed method and it is shown that the proposed system produces better capability in sensing thin crack. Furthermore, Protopapadakis et al. proposed an automatic robotic inspector which monitor tunnel condition in 2016 [19]. In his work, a CNN is employed for visual inspection of the robot. By utilising CNN, high-level discriminative features for complex non-linear pattern classification can be produced. These features later are used to calculate real-time 3D information to identify the crack position and orientation. In addition, studies for crack detection on pavement using CNN have been conducted [20, 21]. In these studies, pavement images are employed to train neural networks. However, unlike [20] which trains the neural networks from scratch, in [21] transfer learning using a pre-trained VGG 16 network is performed. In this way, a pre-trained model is finetuned thus it can be more suitable for the new prediction task. Both studies manage to detect a presence of crack in the pavement images. However, despite some progress have been made in the implementation of deep learning for infrastructure condition monitoring, this implementation is only limited to detection of defect. In addition, in image processing field, quantifying bridge condition from its image still remains a challenging task [22]. Furthermore, the research mentioned previously employed some images which were taken specifically for the research. Some of the research also involved expert in labelling the images. In this study, images are downloaded from crowd and no expert is involved in the process. This can further provide a significant challenge in predicting a bridge's load carrying capacity.

### 2.4. Summary of Novelty and Contribution

This study is primarily aimed at addressing the global development challenges where bridge capacity data is missing and hindering reconstruction. Current methods rely on human expertise through either visual inspection or testing. However, this is expensive, slow, and reliant on the availability of expertise. In this paper, by utilising deep learning as an automatic structure identification method, we have created a cheap, scalable, transparent, and verifiable bridge load identification tool. In time, this method can be standardized and used on smartphones by non-experts. The propose method will require no physical model of a bridge and thus can provide solution for countries that are lacking in civil engineering experts or bridge documentation. In addition, the trained CNN can benefit other researchers through transfer learning.

# 3. Methodology

This research proposes an alternative method for estimation of load carrying capacity of bridges. This method employs crowd sourced bridge images as an input and provide an estimation about either its load rating or design load, as a proxy for the load carrying capacity. The methodology performed in this research is shown in **Figure 1**. The main activities are performed in this project:
1. Data collection of bridge images and load ratings of the bridges
2. Neural network training and testing
3. Evaluation of performance and optimisation

### 3.1. Data Collection and Preparation

In this part of our study, the bridge database of load ratings and corresponding crowd sourced bridge images are collected. For the load ratings, the National Bridge Inventory (NBI) Database published by American Federal Highway Administration (FHWA) [23] was employed. This database provides information about all bridges in America. Some information such as inventory number, location, features, design load, construction, condition, and load rating of bridges can be retrieved from this database. Some of this information was used in labelling bridge images. To obtain crowd sourced images for training the CNN, web scraping was conducted. In this part of the research, an interface program for collecting images from the web was built using Python. By utilising the program, 54458 bridge images from 6753 bridges were collected from a website www.bridgehunter.com [24]. In addition, inventory number and the state ID of these bridges were also collected since they will be used as a matching information with the NBI database. To give identifier on these images, additional ID number was assigned on each bridge. This additional number was used as the folder name for the images. **Figure 2** shows some samples that are collected for this research.

In obtaining either the load rating or design load information for the labelling step of the bridge images, both bridge inventory number and state ID were used as a matching information. State ID was used since bridges with similar inventory number can be found in different states. Hence, by using this method, information about load rating and design load for each image can be obtained from NBI database. However, it is difficult to apply this method in some bridges due to the difference in data format in some states. In this case, the data in the NBI database has to be processed before it can be used as a matching information with the data from bridgehunter database. **Table 1** shows the number of samples according to the availability of the load carrying capacity information. Notice from **Table 1**, there are some samples that have no label. This condition occurs due to the unavailability of either load rating or design load data for the corresponding bridges in NBI database. Due to this problem, some samples can only be used for developing either load rating or design load prediction model.

### 3.2. Network Training and Testing

Instead of training a neural network from scratch, in this research transfer learning method was performed. In this way, a pre-trained network is finetuned in order to suit our application. The implementation of transfer learning using pre-trained network such as AlexNet, VGG-16, and GoogLeNet has shown a great potential in solving wide domain of image classification purpose [14]. It has been shown that, some networks created using transfer learning technique can produce better performance than networks that are built from scratch [14, 15]. In this research AlexNet [9] was utilised as a pre-trained network for transfer learning. AlexNet is a neural network that has been trained to classify 1000 objects. For transfer learning, the neural network is altered according to the number of classes in a dataset we are working with. It is performed by modifying the number of connections at the final fully connected layer to match the number of classes in the dataset. To obtain training and testing data, the original dataset was randomly split into 80% and 20% for training and testing data respectively.

### 3.3. Evaluation of Performance

To evaluate the neural networks performance, some parameters such as accuracy, precision, recall, and F1 score were measured for each prediction model. The precision, recall, and F1 score are given by (1), (2), and (3) respectively [25]:

$$\text{precision} = \frac{TP}{TP+FP} \quad (1)$$

$$\text{recall} = \frac{TP}{TP+FN} \quad (2)$$

$$\text{F1 score} = 2 \cdot \frac{\text{precision} \cdot \text{recall}}{\text{precision}+\text{recall}} \quad (3)$$

Where is TP true positive, FN is false negative, and FP is false positive.

### 3.4. Dataset Modification

In this research, dataset variation and its impact to the prediction performance was conducted. Some variations investigated in this research were the variation in number of classes, bridge images completion, and colour. The performance of the prediction model trained using these datasets was investigated.

#### 3.4.1. Variation in Prediction Class

As it has been mentioned previously, the proposed system utilises a bridge image to provide estimation about either its load rating or design load and this estimation is given in range of loading. This range of loading is defined by prediction classes and classification task is performed by training prediction models. In this part of the research, CNN-based prediction models were trained by varying the number of classes for classification. This was performed to find dataset configuration which produces the highest performance in predicting either load rating or design load of bridges. In addition, the effect of imbalance database was also investigated.

Data distribution of samples according to their load rating can be seen in **Figure 3**. In this research, estimation of load rating was performed by discretising the load rating value to obtain class labels. By using these labels, multiclass classification was performed for the estimation. To create dataset variation for load rating prediction, the class range was modified. In addition, the class interval modification took into account the number of samples obtained in each class. If one class is only formed by small number of images, this class is combined with the class adjacent to this class. As an example, in both dataset LR1 and LR2, due to small number of samples with 0-5 tons of load rating, the 0-5 tons class was merged with the 5-10 tons class. From this modification some datasets were created, and these datasets can be seen in **Table 2**.

As it is shown in **Figure 3**, the data obtained for this research was creating an imbalance dataset. Therefore, for each dataset in **Table 2**, a balance dataset was produced by having a down-sampling process of the majority class in the respective imbalance dataset. These balance and imbalance dataset were used to train prediction model and the impact was observed.

Unlike the labelling process for load rating prediction, the design load information has already been discretised in the NBI database. In developing design load prediction model, several datasets were utilised by varying the number of classes used for the prediction. These datasets are shown in **Table 3**. Dataset A was obtained as the original dataset for design load prediction. The class number was slightly modified from the class number given in [4]. It was done to sort these classes in an ascending order. This dataset consists of 12 classes and as it is shown in table 3, this dataset has imbalance dataset where a lot of samples are obtained for class 2, 3, and 5 and only few samples obtained in other classes especially for class 7, 8, and 11 which only have less than 100 samples. Hence, this dataset was not utilised in the research and other datasets were generated to overcome this problem. In DL1, class 7, 8, 11, and 12 were removed from the dataset due to the limited number of samples. Therefore, this dataset only classified 8 classes. In addition, to deal with data imbalance, in DL2 down-sampling of majority classes was performed. In this dataset, 1000 of samples from each majority classes (category 2, 3, and 5) were randomly picked. Due to this process, this dataset only had 5774 samples.

Unlike DL1 and DL2 that removed samples from class 7, 8, 11, and 12, both DL3 and DL4 combined these samples into one class. Therefore, these datasets were used to classify 9 classes. DL3 utilised all samples while down-sampling process was applied on DL3 to generate DL4 in order to create more balance dataset. Finally, both datasets DL5 and DL6 were created by combining samples from class 5, 6, and 9 into one class. It was performed due to the similar load level applied to these classes. Similar to previous datasets, DL5 had imbalance sample distribution while DL6 was generated as the balance version of DL5.

#### 3.4.2. Design Load Prediction and Load Rating Prediction Models Comparison

In order to create fair comparison between load rating prediction model and design load prediction model, some datasets were generated. In these datasets, the load rating class intervals was modified to match those in the design load. These generated datasets can be seen in **Table 4**. In both DL7 and LR9, all images which can be labelled for either load rating or design load prediction were used. However, it can be seen in the table that there is a discrepancy between the number of images available for load rating prediction and the number of images applicable for design load prediction. In addition, it can be seen that both DL7 and LR9 are imbalance dataset. Therefore, both DL8 and LR10 which have same number of samples in every class were created. In order to reduce a prediction bias toward a majority class in imbalance dataset, both DL9 and LR11 were generated. As it can be seen from table 4, this dataset only has 300 samples on each class due to small number of samples in class 6 available for design load prediction.

### 3.4.3. Image Completion Variation

The data obtained from bridgehunter website [24] contains both images showing a complete view of bridges and images showing incomplete bridge (only shows bridge connection, bridge railing, bridge deck, or bridge column). These variation in the samples could affect the performance of the prediction model therefore in this part of the research this impact was studied. **Figure 4** shows some samples that insufficiently represent a bridge. In order to filter these images, a CNN-based prediction was trained to classify whether or not an image is showing view of a bridge completely. From **Table 1**, it can be seen that the number of images which shows complete bridges is lower than the number of samples which shows incomplete bridge. This can be seen as one of the limitations of this research since these images were obtained from the web thus it is challenging to control the quality of the samples. Some datasets were created: dataset which only contained samples showing complete bridges, dataset which only contained images showing incomplete bridge, and dataset which contained all bridge images. To make a fair comparison, these datasets were configured to have similar number of samples hence down-sampling process was implemented. **Table 5** shows the generated datasets.

In **Table 5**, datasets A (A1, A2, and A3) represent datasets which consist of both images showing complete and incomplete bridges, datasets B (B1, B2, and B3) only include images showing complete bridges, and datasets C (C1, C2, and C3) only consist image with incomplete view of bridge. The number described in each dataset shows data distribution on the datasets. Dataset A1, B1, and C1 were datasets created directly from image filtering process based on the quality of the images. However, as it can be seen in table 6, there is a variation in the number of images inside these datasets. To create a fair comparison, the number of images in dataset A, B, and C had to be equal therefore A2, B2, and C2 were generated. As it can be seen in table 6, these datasets have imbalance sample distribution and due to the equal number of samples in B1 and B2, basically B2 is similar to B1. To prevent prediction bias in imbalance datasets, dataset A3, B3, and C3 were produced. As it is shown in **Table 5**, these datasets have almost equal number of samples in each class.

### 3.4.4. Image Colour Variation

This section was conducted in order to investigate the effect of image colour to the performance of prediction model. The grayscale images were obtained by simply converting colourful images into their grayscale versions. Colourful images are formed by a number of pixels and each colour pixel has combination of RGB colour space. In this research, grayscale conversion was performed by calculating the luminance which is defined by [26]:

$$Y = 0.299R + 0.587G + 0.114B \quad (4)$$

Where Y is the grayscale value, R, G, and B are the red, green, and blue intensity respectively. From these grayscale images, a dataset was created. This dataset was then employed to train a CNN-based prediction model. Finally, comparison between this prediction model and a prediction model created in the previous section was made to observe the effect of image colour to the prediction performance.

## 4. Results and Discussion

### 4.1. Variation in number of classes

#### 4.1.1. Load Rating

**Figure 5** shows the neural network performance for load rating prediction trained using 8 datasets with modification in class interval. From the figure, it can be seen that an increase in the class interval yields a better performance. Maximum accuracy of 68.26% and precision of 60% is achieved on neural network trained using dataset LR7. However, the recall and F1 score produced by this network is slightly lower than those achieved by networks trained using dataset LR5, LR6 and LR8. In the figure, the minimum accuracy is obtained from neural network trained using dataset A. In addition, balancing the data using down-sampling method yields to lower accuracy and precision. This decrease in accuracy from the effect of balance dataset can be seen from **Figure 6**. In the figure, it can be seen that for every imbalanced dataset (LR1, LR3, LR5 and LR7), the corresponding balanced dataset (LR2, LR4, LR6, and LR8) always produced lower accuracy. On the other hand, in term of recall and F1 score no similar trend occurs. However, the number of classes should be taken into account when evaluating the network performance. Although levels of accuracy higher than 60% are achieved on models trained using datasets that only have 3 classes. Therefore, accuracy higher than 60% is only achieved for three-classes prediction.

### 4.1.2. Design Load

**Figure 5** shows the performance of neural networks for design load prediction which are trained using various datasets. In the figure, it can be seen that the for imbalanced dataset scenario, neural network trained using DL5 produces the highest performance. Compared to networks trained using DL1 and DL3, this neural network achieves higher accuracy, precision, recall and F1 score. On the other hand, on balance dataset scenario, neural network trained using DL6 performs best. It can be seen from the higher accuracy, precision, recall, and F1 score that are produced by this neural network compared to those produced by neural networks trained using DL2 and DL4. In **Figure 5**, the effect of balance dataset can also be seen. From the figure, it can be seen that balancing the dataset have negative relation with both accuracy and precision. This can be seen from every pair of balance and imbalance data (DL1-DL2, DL3-DL4, and DL5-DL6). However, it is also shown in **Figure 5** that balancing data can improve both the recall and F1 score.

In order to observe the effect of balance dataset to the neural network's prediction, the prediction made by the networks are investigated. For this purpose, neural networks trained using DL5 and DL6 are investigated. **Figure 6** shows the prediction of samples in every category which is made by neural networks trained using these datasets. From **Figure 6**, it can be seen that although neural networks trained using DL6 produces lower accuracy than those trained using DL5, the bias toward majority classes is minimised. This condition can be seen in **Figure 6**. In the figure, it can be seen that in imbalanced dataset, most prediction for samples from minority class is made toward majority class. On the other hand, for balanced dataset, almost in every category, maximum prediction on one category is made in the true category.

The higher accuracy obtained from neural networks trained using DL5 might be affected by the number of samples in the majority class. Although some misprediction produced in the minority classes, the number of samples in majority class is much larger than the number of samples in minority classes. Therefore, the true prediction made in the majority class contributes more than the misprediction hence higher accuracy can be achieved. In addition, unlike misprediction that is produced toward majority class in imbalanced dataset scenario, from the figure it can be seen that for balanced dataset, misprediction mostly occurs to the adjacent category. In this case, the prediction does not deviate too much from the expected category. There are some exceptions such as for samples in 15, 20, and 45 tons category. This might occur due to the data distribution. Although down-sampling has been performed to balance the dataset, the number of samples in the down-sampled category is still twice as much as the number of samples in some categories such as the 27 to, 36 tons with military inclusion, and 45 tons categories. Therefore, the bias prediction toward majority classes is still occurred.

### 4.1.3. Comparison Between Models for Design Load Prediction and Models for Load Rating Prediction

The performance of prediction models can be seen in **Figure 5**. In the figure, it can be seen that for equal number of samples in the dataset, the models created for design load prediction perform better compared to models trained for load rating prediction. Comparisons are made between DL7 and LR9 (case 1), between DL8 and LR10 (case 2), as well as between DL9 and LR11 (case 3).

In case 1, all images available for either load rating prediction or design load prediction are utilised. In **Figure 5**, it can be seen that DL7 produce higher accuracy and precision compared to LR9 and significant differences on these parameters are produced between these models. However, in case 1 the load rating prediction model produces slightly higher recall which leads to a slightly higher F1 score compared to the design load prediction model.

With equal number of images in each class (case 2), it can be seen that the design load prediction model produces higher performance than the load rating prediction model. This can be seen in **Figure 5** from the higher accuracy, precision, recall, and F1 score obtained from the design load prediction model. Another interesting feature that can be observed in this figure is the higher accuracy achieved by the load rating prediction model compared to its performance on case 1. This event might occur due to the more imbalance dataset used in case 2 compared to the dataset for case 1 which can be seen in table 5. In imbalance dataset scenario, more predictions tend to be made on a majority class which can lead to the increase in accuracy. However, the increase in accuracy doesn't mean a better performance since in case 2, the load rating prediction model produce lower precision, recall, and F1 score.

In balance datasets scenario (case 3), it can be seen that the design load prediction model achieves higher accuracy, precision, recall, and F1 score compared to the load rating prediction model. In this case, with similar number of data and minimum bias from imbalance dataset, the design load prediction model still manages to perform better than the load rating prediction model. Note that lower performance is produced in this scenario due to the small number of samples utilised in the datasets for case 3.

In this section, it has been shown that predicting design load from bridge images is more achievable compared to predicting load rating from bridge images. Load rating of a bridge is affected by its condition. However, it is a challenging task to quantify a bridge condition from its image. This situation might introduce error in the samples. On the other hand, bridge condition provides no effect on the capacity which the bridge is designed for. This might lead to a better performance in design load prediction model.

### 4.2. The Impact of Image Completion to the Prediction Performance

The performance of the prediction models produced by datasets with various image quality on three different scenarios can be seen from **Figure 5**. For case 1, it can be seen that the neural networks trained using dataset consisting images which represent a bridge (B1) perform the best among all. This can be seen from the highest accuracy, precision, recall, and f1 score produce by this prediction model compared to the other models. Bear in mind that this performance is achieved using fewer number of samples in the dataset compared to other datasets. The number of data samples used in the training process might explain the slight difference between the accuracy achieved by model trained using dataset A1 and model trained using dataset B1. In this case dataset A1 has more images compared to dataset B1.

In case 2 where all datasets have equal number of data samples in every class, it can be seen that prediction models trained using good quality images produce the highest performance among all. Furthermore, since all datasets have equal number of data samples, significant difference in the models' performance can be identified in this scenario. In addition, it is also shown that by using equal number of data samples, the model trained using low quality images achieves the lowest performance in term of accuracy, precision, recall, and F1 score.

In case 3 which is balance datasets scenario, it can be seen that the model trained using good also achieves the highest accuracy, precision, recall, and F1 score among all prediction models. Similar with the previous section, decreases in performance found in these prediction models. This might be due to the decrease of images number that are used to train these models because of down-sampling process used to create balance datasets.

In this comparison, the result shows that in all 3 cases the models trained using good quality data produce the highest performance. From this comparison, it can be concluded that to obtain satisfactory performance, image quality plays an important role. This can be one factor that limits this research since it is challenging to obtain images with the right angle from web scraping.

### 4.3. The Impact of Image Colour to the Prediction Performance

Performances of prediction model trained using colourful images and the performance of grayscale trained prediction model are provided in **Figure 5**. It can be seen that the implementation of grayscale images can worsen the performance of the load rating or design load estimation model. All parameters achieved by the model trained using grayscale image are lower than the parameters of prediction model trained using colourful images. This can be described by the effect of colour in detecting a material where colourful image can give more information about the material that forms an object. Hence, it is more suitable to use colourful images for this kind of application.

### 4.4. Proposed Optimisation Method for Performance Improvement

The result from model prediction that has been presented show unsatisfactory performance which can be inferred from the low accuracy. In order to analyse this condition, a neural network for design load prediction trained using DL6 is taken as a case study. **Figure 7** describes the training and validation process on a neural network trained using this dataset. From the figure, it can be seen that although the training accuracy reach 97% during the training process, overfitting occurs where the validation accuracy is not increasing after several iterations. Improvement of validation accuracy only occurs in the first 500 iteration before fluctuating validation accuracy is achieved. This leads to a significant difference between training and testing accuracy of the prediction model. This condition happens on all neural networks. Due to this situation, early stopping can be implemented to stop the training process whenever no increase in validation accuracy is achieved. However, no significant improvement is obtained only by using early stopping method. As it can be seen in **Figure 7**, early stopping can only slightly improve the accuracy into 41%, which is only 3% improvement from 38% produced without early stopping.

In order to observe the prediction error made by the model, an error probability distribution from the prediction is created. This error probability distribution is shown in **Figure 8**. From the figure, it can be seen that the error created follows normal distribution. In this case, most predictions are made into the right class and most error occurs when predicting a bridge into a class adjacent to the right class. Both load rating and design load prediction produces similar trend in the error distribution.

From this error distribution, it is possible to increase the performance by merging two classes adjacent to each other. The simplest way is by converting the multiclass classification into binary classification. In this case, the model is not predicting the exact value of either load rating or design of a bridge in the image. However, the model is used to predict whether a bridge load rating or design load is higher or lower than a certain value. This conversion can be seen in **Figure 9**.

Using binary classification conversion, the network is now used to predict:
- If a bridge has design load lower than 10 ton (level 1)
- If a bridge has design load lower than 15 ton (level 2)
- If a bridge has design load lower than 20 ton (level 3)
- If a bridge has design load lower than 27 ton (level 4)
- If a bridge has design load lower than 36 ton (level 5)

Notice from **Figure 9**, there are some regions that becomes either true positive or true negative after conversion. By performing this conversion, improvement on prediction model can be obtained. The accuracy, precision, recall, and F1 score produced by the system on each low level after conversion can be seen in **Figure 5**. Following the conversion, it can be seen in **Figure 5** that an increase in the performance is produced. In addition, it is shown that the maximum accuracy, precision, recall, and F1 score are produced when predicting in level 5. In this level the accuracy, precision, recall, and F1 score are 90.89, 95.21, 94.99, and 95.10 respectively.

## 5. Conclusion

In this research, a novel neural network for estimating the load rating and design load of bridges have been trained based on crowd sourced image data. By using equal number of samples, the neural networks trained for design load prediction produce higher performance than the neural network for load rating prediction. This can be seen from the higher accuracy, recall, precision and F1 score of the models trained for design load prediction compared to those of the load rating prediction models. The image quality affects the performance of prediction models. Therefore, in order to improve the performance, the quality of images used to train neural networks should be taken into account. It was found that the used of colourful images for this application is more suitable than using grayscale images. Furthermore, converting multiclass classification into binary classification can improve the prediction performance. Our future work will focus on refining the training process by artificially generating bridge data and images.

**Acknowledgments**
N/A

**Ethical Statement**
No ethical statement is required for this study.

**Funding Statement**

This research was funded by Indonesian Endowment Fund for Education (PRJ-589 /LPDP.3/2017). The paper is co-funded by: Lloyd's Register Foundation's Data-Centric Engineering Program at the Alan Turing Institute (EP/N510129/1).

**Data Accessibility**
Our data are available from Dryad Digital Repository:
https://datadryad.org/review?doi=doi:10.5061/dryad.6br51tn

**Competing Interests**
*We have no competing interests.*

**Authors' Contributions**
A.P. performed data collection, training and testing prediction model, and wrote the manuscript. I.L and W.G created concept of the study, designed the study, coordinated the study, and helped producing the manuscript, A.K. helped write the manuscript. All authors gave final approval for the publication of this paper.

Table and figure captions should be included at the end of the manuscript file and should be brief and informative.

**Table 1 Images obtained for load rating and design load prediction**

| Remark | Number of images | Images showing a full picture of a bridge | Number of images only shows part of a bridge |
|---|---|---|---|
| All Data | 54458 | N/A | N/A |
| Images for Design Load Estimation | 18821 | 7837 | 10984 |
| Images for Load Rating Estimation | 43180 | 13510 | 29670 |

**Table 2 Datasets for CNN training and Testing**

| Dataset | Total Number of Classes | Class Description |
|---|---|---|
| LR1 and LR2 | 7 | 1. 0-10 tons<br>2. 10-15 tons<br>3. 15-20 tons<br>4. 20-25 tons<br>5. 25-30 tons<br>6. 30-35 tons<br>7. >35 tons |
| LR3 and LR4 | 5 | 1. 0-10 tons<br>2. 10-20 tons<br>3. 20-30 tons<br>4. 30-40 tons<br>5. >40 tons |
| LR5 and LR6 | 3 | 1. 0-15 tons<br>2. 15-30 tons<br>3. >30 tons |
| LR7 and LR8 | 3 | 1. 0-20 tons<br>2. 20-40 tons<br>3. >40 tons |

**Table 3 Data distribution for design load prediction**

| Class | Metric | Remark | Dataset A | DL1 | DL2 | DL3 | DL4 | DL5 | DL6 |
|---|---|---|---|---|---|---|---|---|---|
| 1 | H10 | 10 ton | 928 | 928 | 928 | 928 | 928 | 928 | 928 |
| 2 | H15 | 15 ton (3000 front 12000 rear) | 4674 | 4674 | 1000 | 4674 | 1000 | 4674 | 1000 |
| 3 | H20 | 20 ton (4000 front 16000 rear) | 1913 | 1913 | 1000 | 1913 | 1000 | 1913 | 1000 |
| 4 | HS15 | 27 ton (3000 front 12000 mid & rear) | 460 | 460 | 460 | 460 | 460 | 460 | 460 |
| 5 | HS20 | 36 ton (4000 front 16000 mid & rear) | 3991 | 3991 | 1000 | 3991 | 1000 | 5067 | 1000 |
| 6 | HS20+mod | Equal to HS20 with the inclusion of military loading | 491 | 491 | 491 | 491 | 491 | 0 | 0 |
| 7 | pedestrian |  | 3 | 0 | 0 | 0 | 0 | 0 | 0 |
| 8 | railroad |  | 56 | 0 | 0 | 0 | 0 | 0 | 0 |
| 9 | HL93 | Equal to HS20 with an addition of road calculation | 585 | 585 | 585 | 585 | 585 | 0 | 0 |
| 10 | HS25 | 45 ton or greater | 310 | 310 | 310 | 310 | 310 | 332 | 332 |
| 11 | >HL93 | Greater than HL93 | 22 | 0 | 0 |  | 0 | 0 | 0 |
| 12 | other | For other bridge which employs other standard than AASHTO | 107 | 0 | 0 | 188 | 188 | 0 | 0 |
| Total |  |  | 13540 | 13352 | 5774 | 13540 | 5962 | 13540 | 5962 |

**Table 4 Generated datasets for comparison of load rating and design load prediction models**

| Class | Number of samples |
|---|---|

| | DL7 | LR9 | Both DL8 and LR10 | Both DL9 and LR11 |
|---|---|---|---|---|
| 1 | 496 | 2098 | 496 | 300 |
| 2 | 2496 | 1153 | 1153 | 300 |
| 3 | 1201 | 1358 | 1201 | 300 |
| 4 | 287 | 2483 | 287 | 287 |
| 5 | 3171 | 2802 | 2802 | 300 |
| 6 | 186 | 920 | 186 | 186 |

**Table 5 Datasets for investigation of image completion to the performance of prediction model**

| Class | DL10 | DL11 | DL12 | DL13 | DL14 | DL15 | DL16 | DL17 | DL18 |
|---|---|---|---|---|---|---|---|---|---|
| 1 | 1456 | 496 | 300 | 496 | 496 | 300 | 960 | 496 | 300 |
| 2 | 6389 | 2496 | 300 | 2496 | 2496 | 300 | 3893 | 2496 | 300 |
| 3 | 2635 | 1201 | 300 | 1201 | 1201 | 300 | 1434 | 1201 | 300 |
| 4 | 706 | 287 | 287 | 287 | 287 | 287 | 419 | 287 | 287 |
| 5 | 5406 | 2460 | 300 | 2460 | 2460 | 300 | 2946 | 2460 | 300 |
| 6 | 641 | 312 | 300 | 312 | 312 | 300 | 329 | 312 | 300 |
| 7 | 962 | 399 | 300 | 399 | 399 | 300 | 563 | 399 | 300 |
| 8 | 379 | 186 | 186 | 186 | 186 | 186 | 193 | 186 | 186 |

**Figures**

For final submissions, figures should be uploaded as separate files.

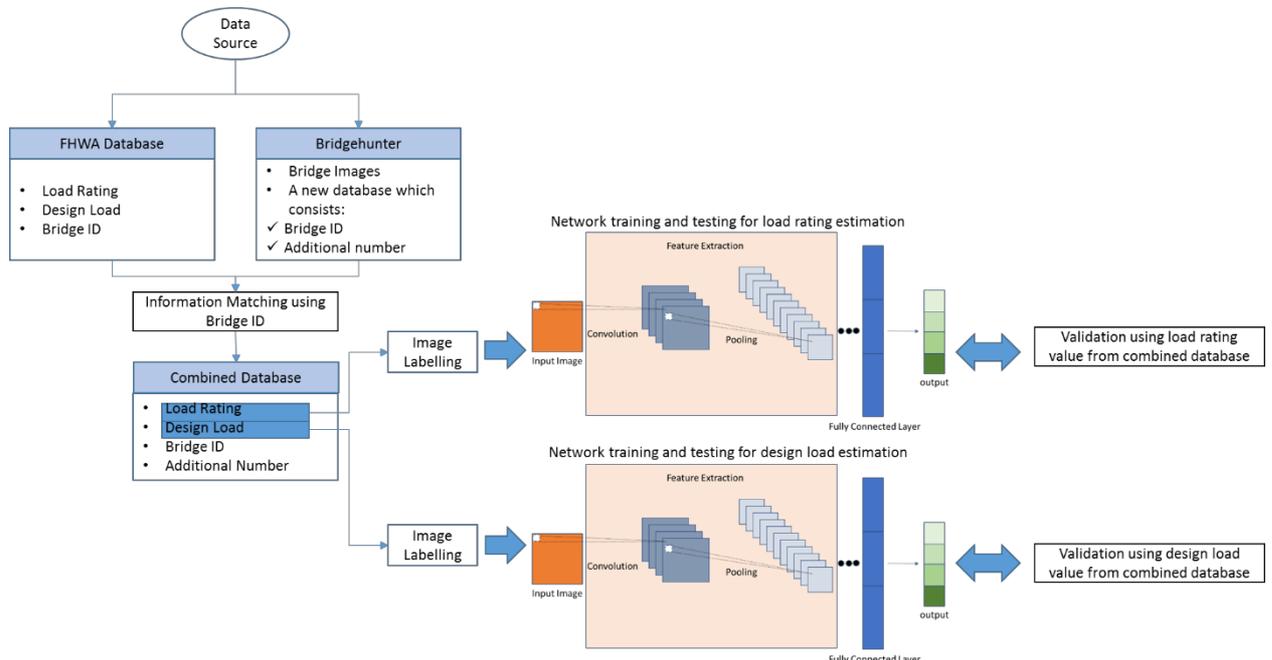

**Figure 1 Schematic of data collection, data labelling, network testing and training, and evaluation**

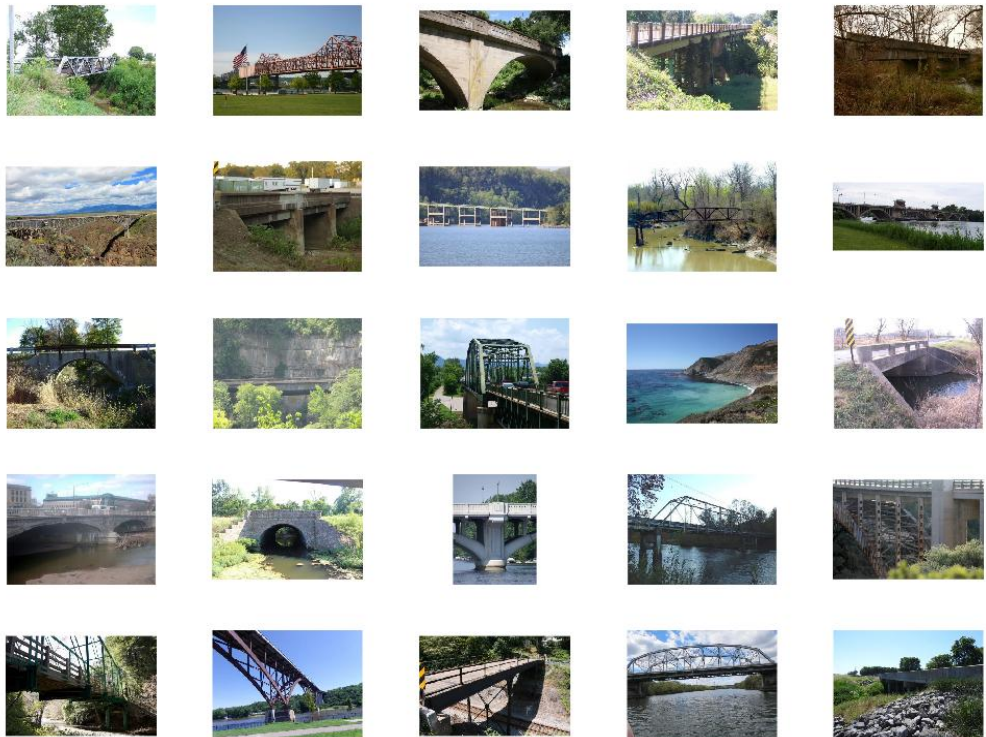

**Figure 2 Images Collected from Web Scraping**

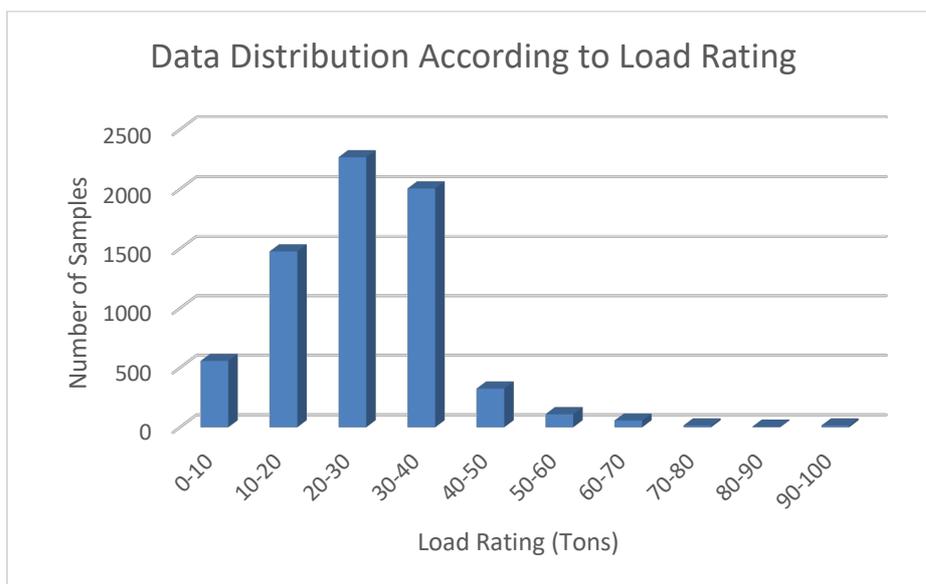

**Figure 3 Data distribution according to the load rating**

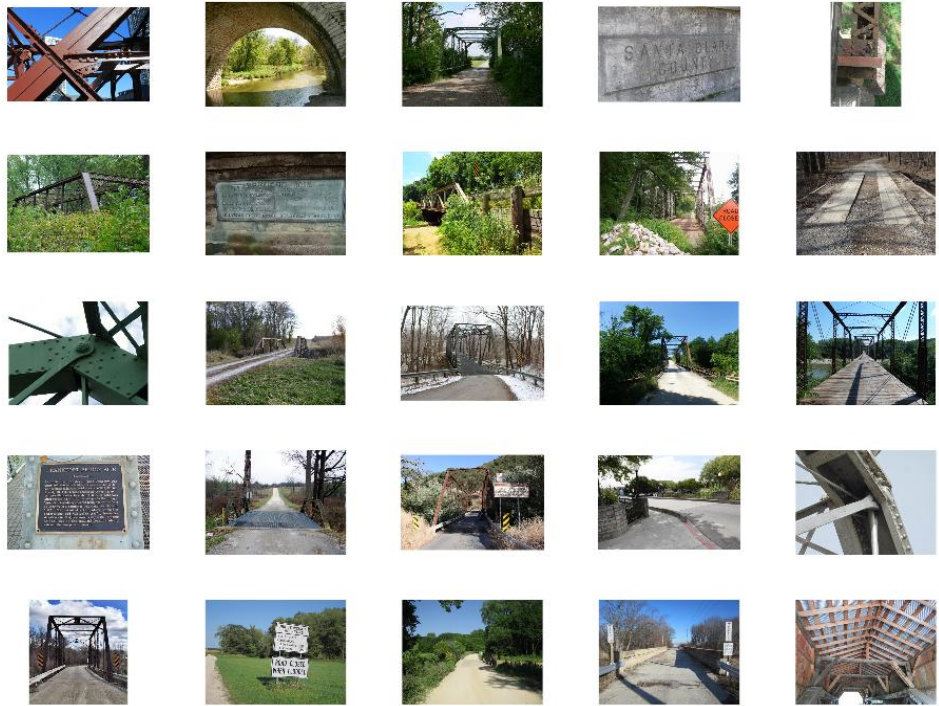

Figure 4 Images showing only part of a bridge

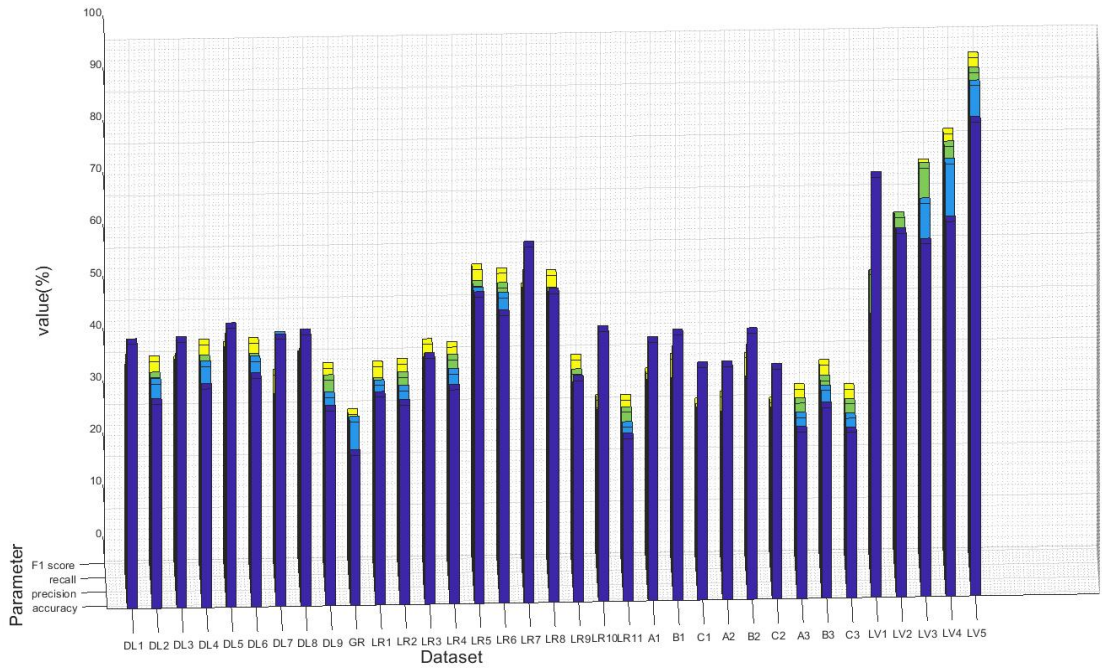

Figure 5 Performance of model trained for load rating or design load prediction

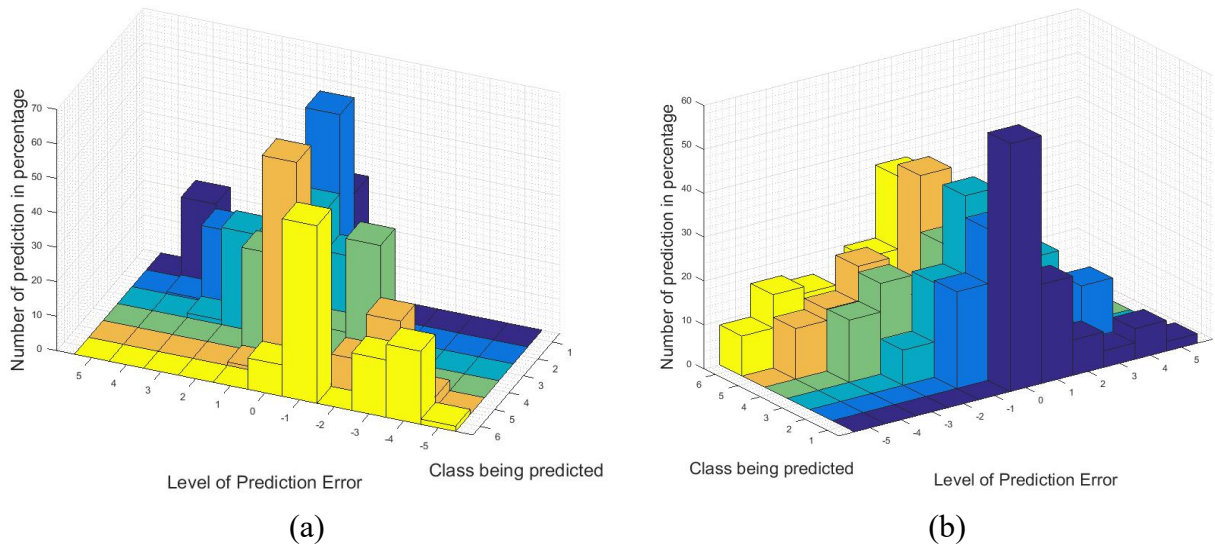

(a)　　　　　　　　　　　　　　　(b)

**Figure 6 Distribution of error made by neural networks trained using dataset DL5 (imbalanced dataset) and DL6 (balanced dataset). Minus (–) sign indicates the prediction is lower than actual class. The value determine how far it deviates from the actual class.**

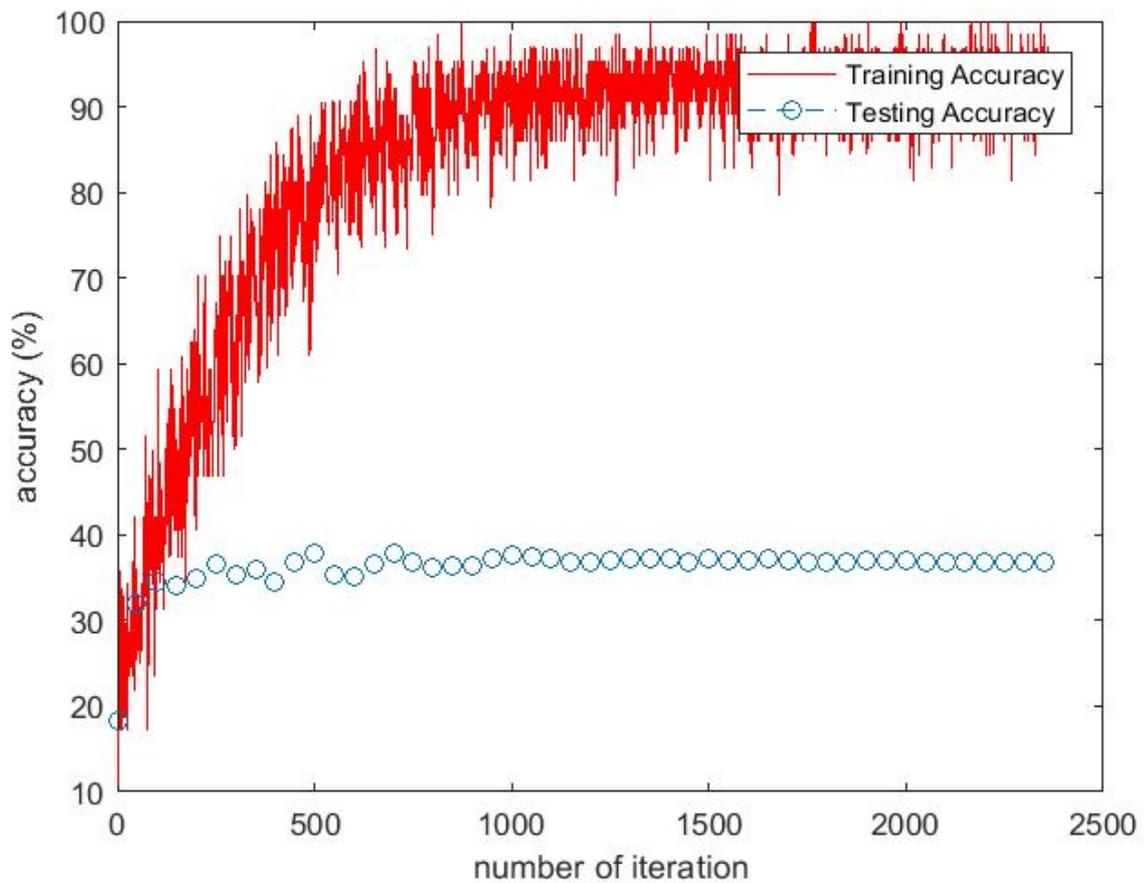

**Figure 7 Training and validation accuracy of neural network for load rating prediction**

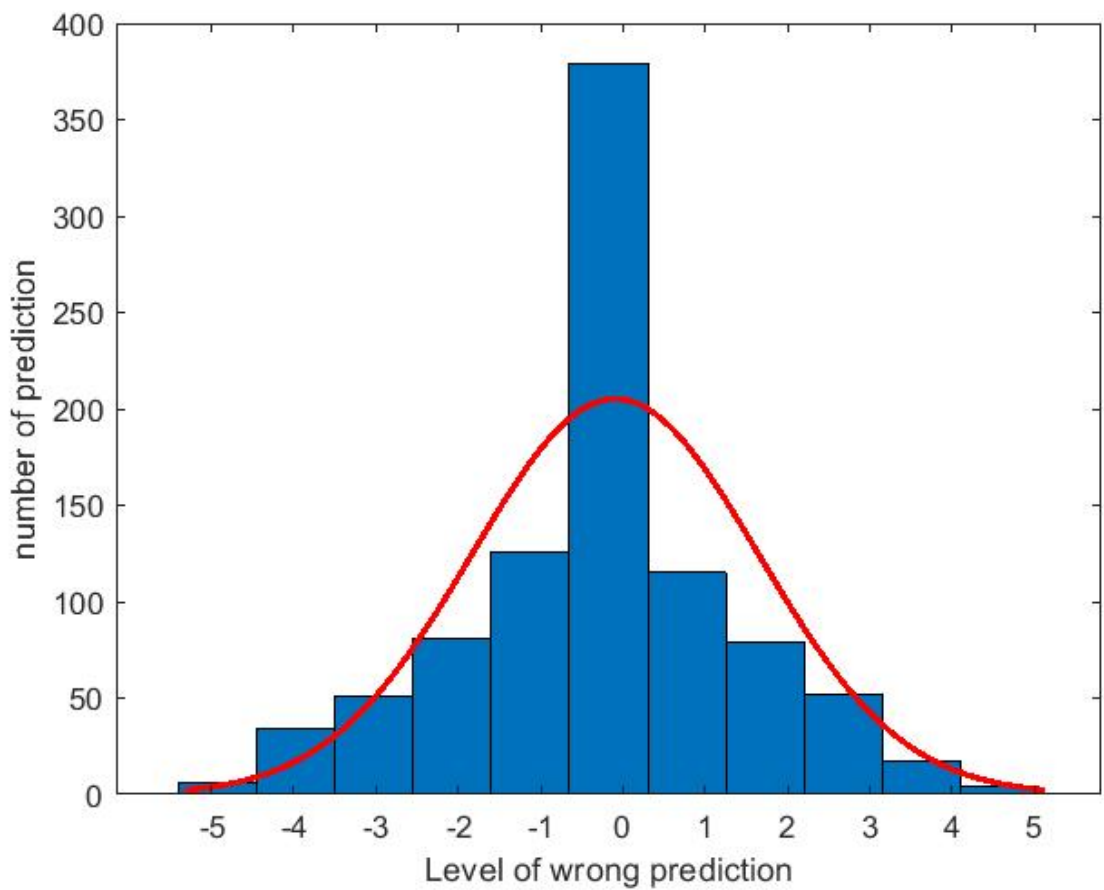

**Figure 8 Error probability from design load prediction**

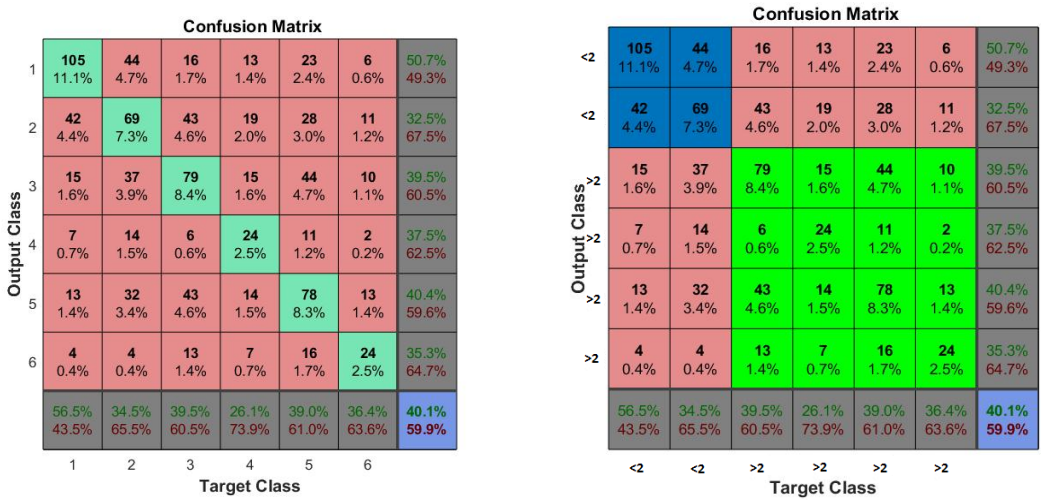

**Figure 9 Conversion from multiclass classification into binary classification to detect if a bridge load rating is lower than 15 ton. After conversion, both true positive and true negative region become larger thus enhancing the prediction performance.**